\documentclass[letterpaper]{article} 
\usepackage[draft]{aaai25}  
\usepackage{times}  
\usepackage{helvet}  
\usepackage{courier}  
\usepackage[hyphens]{url}  
\usepackage{graphicx} 
\usepackage{amsmath} 
\usepackage{amsfonts}
\usepackage{natbib}  
\usepackage{caption} 
\usepackage{algorithm}
\usepackage{xcolor}
\usepackage{algorithmic}

\usepackage{booktabs} 
\usepackage{multirow}
\usepackage{graphicx}
\usepackage{subcaption}

\usepackage{newfloat}
\usepackage{listings}
\DeclareCaptionStyle{ruled}{labelfont=normalfont,labelsep=colon,strut=off} 
\floatstyle{ruled}
\newfloat{listing}{tb}{lst}{}
\floatname{listing}{Listing}
\title{G3PT: Unleash the power of Autoregressive Modeling in 3D Generation via Cross-scale Querying Transformer}

\author {
    Jinzhi Zhang,
    Feng Xiong*,
    Mu Xu
}
\affiliations {
    AMAP\\
    {\{wushou.zjz, xf250971, xumu.xm\}@alibaba-inc.com~}
}

\usepackage{bibentry}

\usepackage{xcolor}

\begin{document}

\maketitle 
\footnotetext{*~Equal contribution.}

\begin{abstract}

Autoregressive transformers have revolutionized generative models in language processing and shown substantial promise in image and video generation.
However, these models face significant challenges when extended to 3D generation tasks due to their reliance on next-token prediction to learn token sequences, which is incompatible with the unordered nature of 3D data.
Instead of imposing an artificial order on 3D data, in this paper, we introduce G3PT, a scalable coarse-to-fine 3D generative model utilizing a cross-scale querying transformer. The key is to map point-based 3D data into discrete tokens with different levels of detail, naturally establishing a sequential relationship between different levels suitable for autoregressive modeling. Additionally, the cross-scale querying transformer connects tokens globally across different levels of detail without requiring an ordered sequence.
Benefiting from this approach, G3PT features a versatile 3D generation pipeline that effortlessly supports diverse conditional structures, enabling the generation of 3D shapes from various types of conditions.
%
Extensive experiments demonstrate that G3PT achieves superior generation quality and generalization ability compared to previous 3D generation methods. Most importantly, for the first time in 3D generation, scaling up G3PT reveals distinct power-law scaling behaviors.

\end{abstract}

\section{Introduction}

In recent years, the field of 3D shape generation has experienced significant advancements. One notable approach is the use of Large Reconstruction Models (LRMs) ~\cite{hong2023lrm, tochilkin2024triposr}, which convert images into 3D shapes through a pipeline that employs transformers~\cite{c:22} to create and optimize implicit 3D representations with multi-view image supervision. Another approach extends 2D diffusion models~\cite{rombach2022high} into the 3D domain, aiming to combine multi-view images into cohesive 3D shapes using techniques such as sparse view reconstruction~\cite{li2023instant3d, ji2020surfacenet_plus} and score distillation sampling~\cite{poole2022dreamfusion}. However, these methods heavily depend on the fidelity of the multi-view images and often struggle to generate high-quality meshes, particularly when capturing intricate geometric details.
%
{To address these challenges, a newer paradigm~\cite{zhang20233dshape2vecset} leverages 3D variational auto-encoders to compress high-resolution point clouds into a compact latent space before performing diffusion to directly generate 3D shapes.}
Despite its potential, this approach is limited by lengthy training times and the lack of a guided scaling strategy, which constrain its effectiveness and scalability.


\begin{figure*}[ht]
\centering
\includegraphics[width=0.85\textwidth]{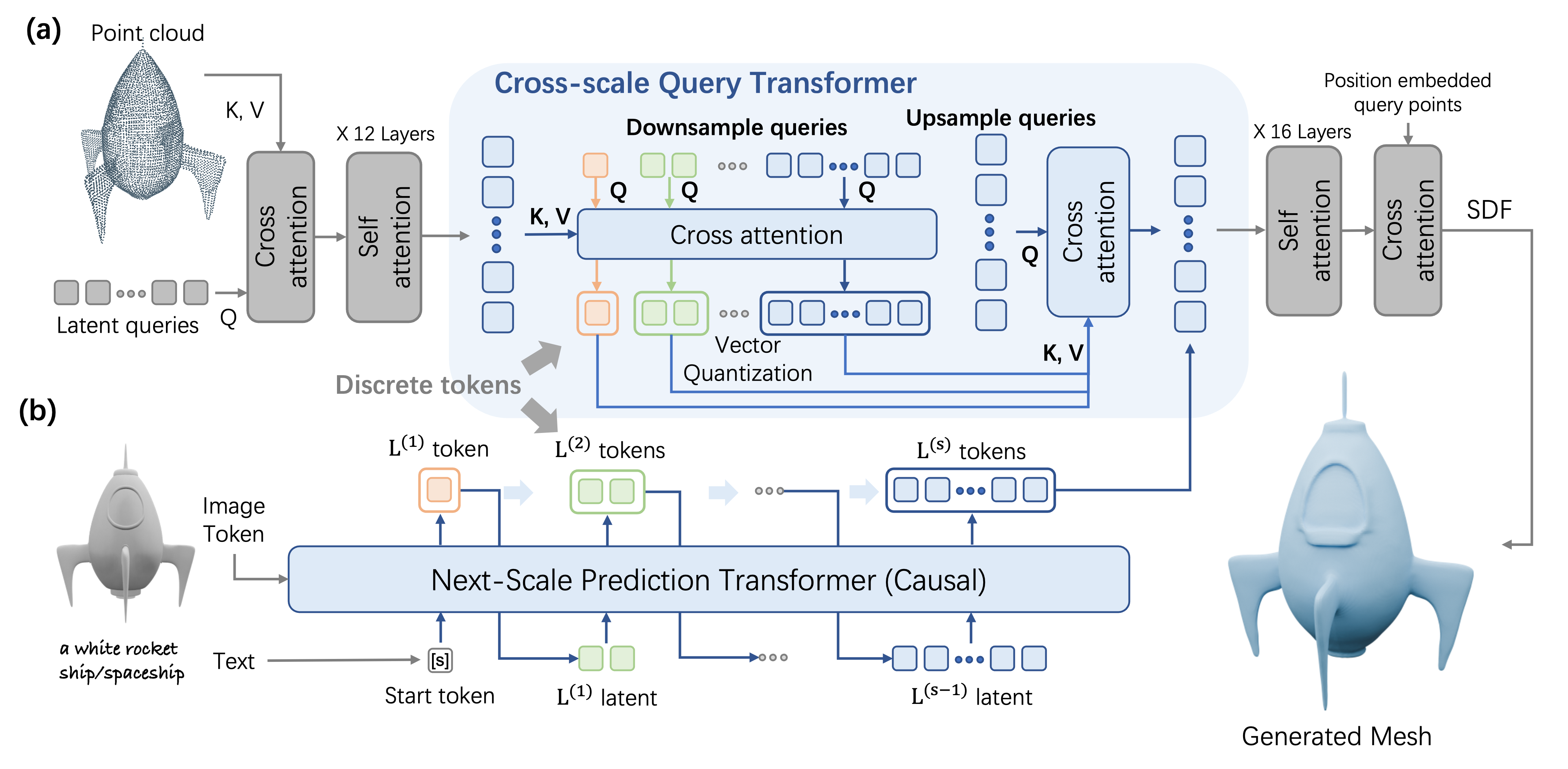} 
\vspace{-8pt}
\caption{Overall pipeline for processing and generating unordered 3D data. (a) G3PT starts by encoding the input point cloud into discrete scales of token maps, each representing different levels of detail. The proposed Cross-scale Querying Transformer (CQT) utilizes a cross-attention layer with varying numbers of queries to globally connect tokens across different scales, without requiring the tokens to be in a specific order. The final output is the SDF value for each query point.
(b) CQT enables 3D generation from coarse to fine scales under various conditions. An autoregressive transformer is trained using next-scale prediction.}
\vspace{-10pt}
\label{fig: pipeline}
\end{figure*}

In parallel, the emergence of AutoRegressive (AR) Large Language Models~\cite{Brown2020LanguageMA} and multimodal AR models~\cite{liu2023llava} has ushered in a new era in artificial intelligence. These models demonstrate exceptional scalability, generality, versatility, and multimodal capabilities. At the core of these AR models is the tokenizer~\cite{esser2021taming}, which transforms diverse data into discrete tokens, enabling the model to employ self-supervised learning for next-token prediction.

AR models have also made notable advancements in visual generation, leveraging their sequential processing capabilities to construct images as grids of 2D tokens, which are then flattened into 1D sequences using a raster-scan process~\cite{LFQ}. However, when extended to 3D generation tasks, these models encounter significant challenges due to their reliance on next-token prediction, which is incompatible with the unordered nature of 3D data. Consequently, current attempts to use structured 3D volumes~\cite{cheng2023sdfusion} or 2D triplanes~\cite{wu2024direct3d} to represent 3D features struggle to learn effective feature sequences from unordered 3D data. Similarly, approaches like MeshGPT~\cite{siddiqui2024meshgpt} tokenize serialized mesh data with a GNN-based encoder~\cite{Zhou2018GraphNN} but still require manually defined sequences, such as z-ordering~\cite{wu2024ptv3}, 
{limiting their effectiveness to complex datasets.}
Conversely, 3D data inherently exhibits level-of-detail characteristics, with a natural sequential relationship across different scales—a concept well established in 3D rendering~\cite{lindstrom1996real} and reconstruction~\cite{zhang2021surrf}. Building on this insight, we introduce G3PT, a scalable coarse-to-fine 3D generative model that effectively maps point-based 3D data into discrete tokens at various levels of detail, creating a sequential relationship ideally suited for autoregressive modeling.
Unlike recent Visual Autoregressive~\cite{VAR} models, which also use "next-scale prediction" but rely on average pooling and bilinear interpolation—methods poorly suited for unordered data—G3PT employs Cross-scale Querying Transformer (CQT), which uses a cross-attention layer to connect tokens across different scales, enabling global integration of information without imposing a specific token order.

In practice, G3PT includes a transformer-based tokenizer to encode high-resolution point clouds into latent feature maps and decode them into 3D occupancy grids through querying points~\cite{zhang20233dshape2vecset}. 
{During quantization, the cross-scale queries compress latent features into discrete tokens at various scales, creating level-of-detail representations. 
These cross-scale tokens are subsequently decoded by employing the upsample queries, which facilitate another cross-attention to align with the latent features at the corresponding scale.}
With these tokens across different scales, the autoregressive process of G3PT begins at the coarsest scale with only one token, and the transformer predicts the next-scale token map conditioned on all previous ones.
This approach provides G3PT with a versatile 3D generation pipeline, seamlessly supporting diverse conditional structures, including image-based and text-based inputs.
Extensive experiments show that G3PT not only surpasses previous LRM-based and 3D generation methods in terms of generation quality but also, for the first time in 3D generation, reveals distinct scaling-law behaviors.

In summary, the key contributions of this work are:
\begin{itemize}
 \item The introduction of the first cross-scale AutoRegressive modeling framework for generating unordered data, offering insights into AR algorithm for new tasks.
 \item The development of a Cross-scale Querying Transformer that tokenizes 3D data into discrete tokens at varying scales, enabling sequential coarse-to-fine AR modeling.
 \item Demonstration through extensive experiments that G3PT sets a new state-of-the-art in 3D content creation, outperforming previous LRM and diffusion-based methods.
\end{itemize}

\section{Related Work}

\textbf{Large Reconstruction Models: } Extensive 3D datasets~\cite{deitke2023objaverse, deitke2024objaverse} have enabled the development of LRM~\cite{hong2023lrm, tochilkin2024triposr}, utilizing transformers to map image tokens to implicit 3D triplanes with multi-view supervision. Instant3D~\cite{li2023instant3d} and MeshLRM~\cite{wei2024meshlrm} extend LRM from single-view to sparse multi-view inputs by integrating a multi-view diffusion model. Methods like InstantMesh~\cite{xu2024instantmesh} and CRM~\cite{wang2024crm} incorporate Flexicubes~\cite{shen2023flexible} for direct mesh optimization, enhancing smoothness and detail. To improve rendering efficiency, LGM~\cite{tang2024lgm} and GRM~\cite{xu2024grm} replace triplane NeRF with 3D Gaussians~\cite{kerbl20233d}. However, these approaches often prioritize minimizing rendering loss over explicit mesh generation, resulting in coarse or noisy geometry.

\textbf{3D Native Generative Models: } Generating 3D content with direct 3D supervision offers a more efficient approach, yet training 3D generative models directly on 3D data poses significant challenges due to high memory and computational demands. Recent methods, such as MeshGPT~\cite{siddiqui2024meshgpt}, Shap-E~\cite{jun2023shap}, and others~\cite{zhang20233dshape2vecset, zhao2024michelangelo, li2024craftsman, zhang2024clay}, compress 3D shapes into a compact latent space before performing diffusion or autoregressive processes. While MeshGPT shows promise, its performance is limited by the mesh tokenizer. Direct3D~\cite{wu2024direct3d} and LAM3D~\cite{cui2024lam3d} further enhance generation quality by introducing explicit 3D triplane representations. However, extended training cycles and unguided scaling strategies still constrain the efficiency of 3D generation.

\textbf{Autoregressive Models for Image Generation: } Autoregressive models have revolutionized visual generation by sequentially creating images using discrete tokens, produced by image tokenizers~\cite{van2017neural, esser2021taming}. Models like DALL-E~\cite{ramesh2021zero}, RQ-Transformer~\cite{lee2022autoregressive}, and Parti~\cite{yu2022scaling} rely on raster-scan sequences for "next token prediction" within a given scale. VAR~\cite{tian2024visual} introduces a novel "next-scale prediction" approach, which better preserves spatial locality and reduces computational costs. This paper explores the potential of next-scale autoregressive modeling in 3D generation, building on the scalability advantages demonstrated in 2D applications.

\section{Method}

\begin{figure}[t]
\centering
\includegraphics[width=0.5\textwidth]{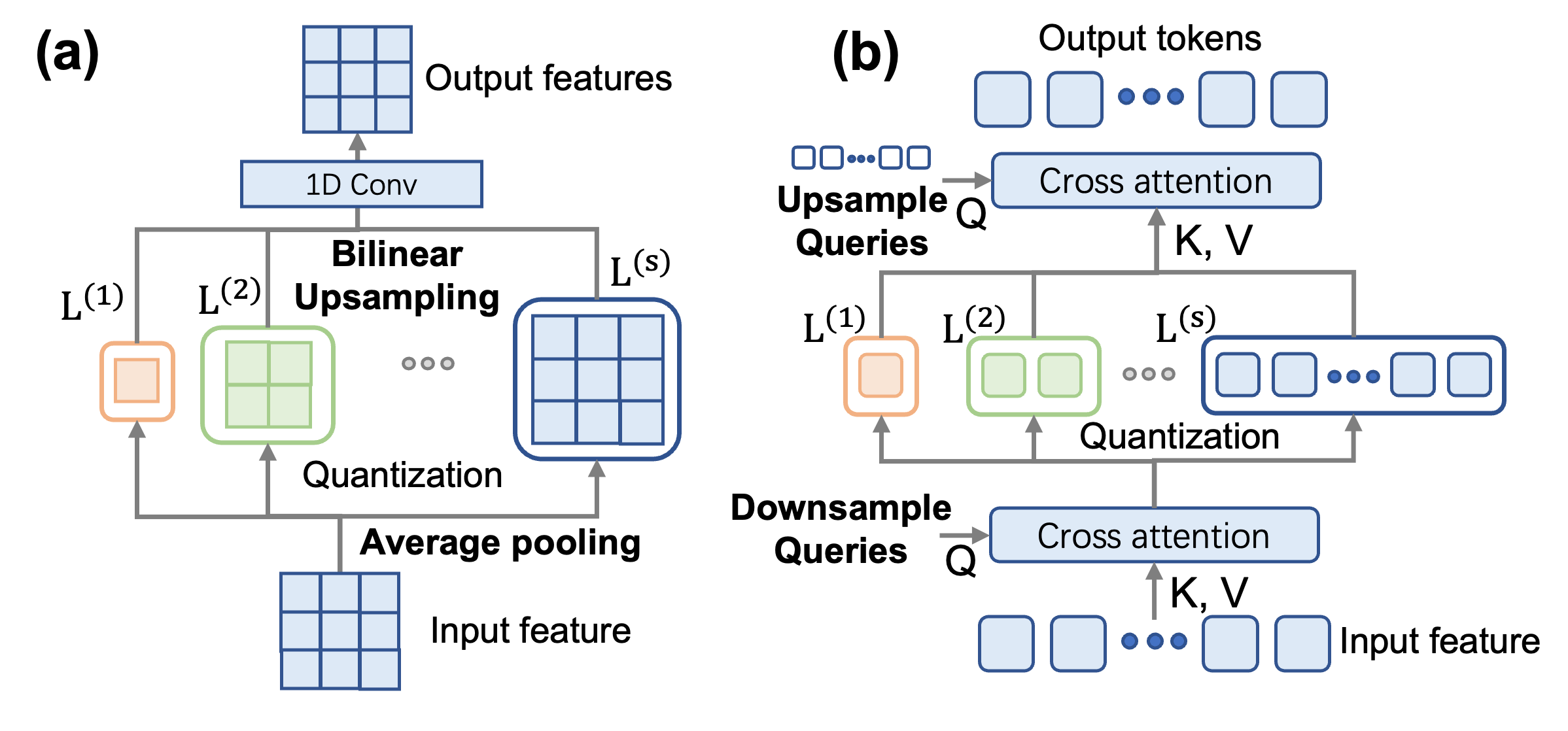} 
\caption{(a) The previous quantization method in VAR~\cite{VAR} relies on average pooling and bilinear upsampling, which are not suitable  for unordered data. (b) Our Cross-scale Vector Quantization (CVQ) overcomes this limitation by using a set of cross-scale learnable latent queries to globally ``pool" and ``upsample" unordered tokens. During the quantization stages, these learnable queries ``downsample" features into fewer tokens at each scale, forming level-of-detail representations. These tokens are then ``upsample"  to their original scale using another cross-attention layer. }
\label{fig: CVQ}
\vspace{-8pt}
\end{figure}

\begin{figure}[t]
\centering
\includegraphics[width=0.5\textwidth]{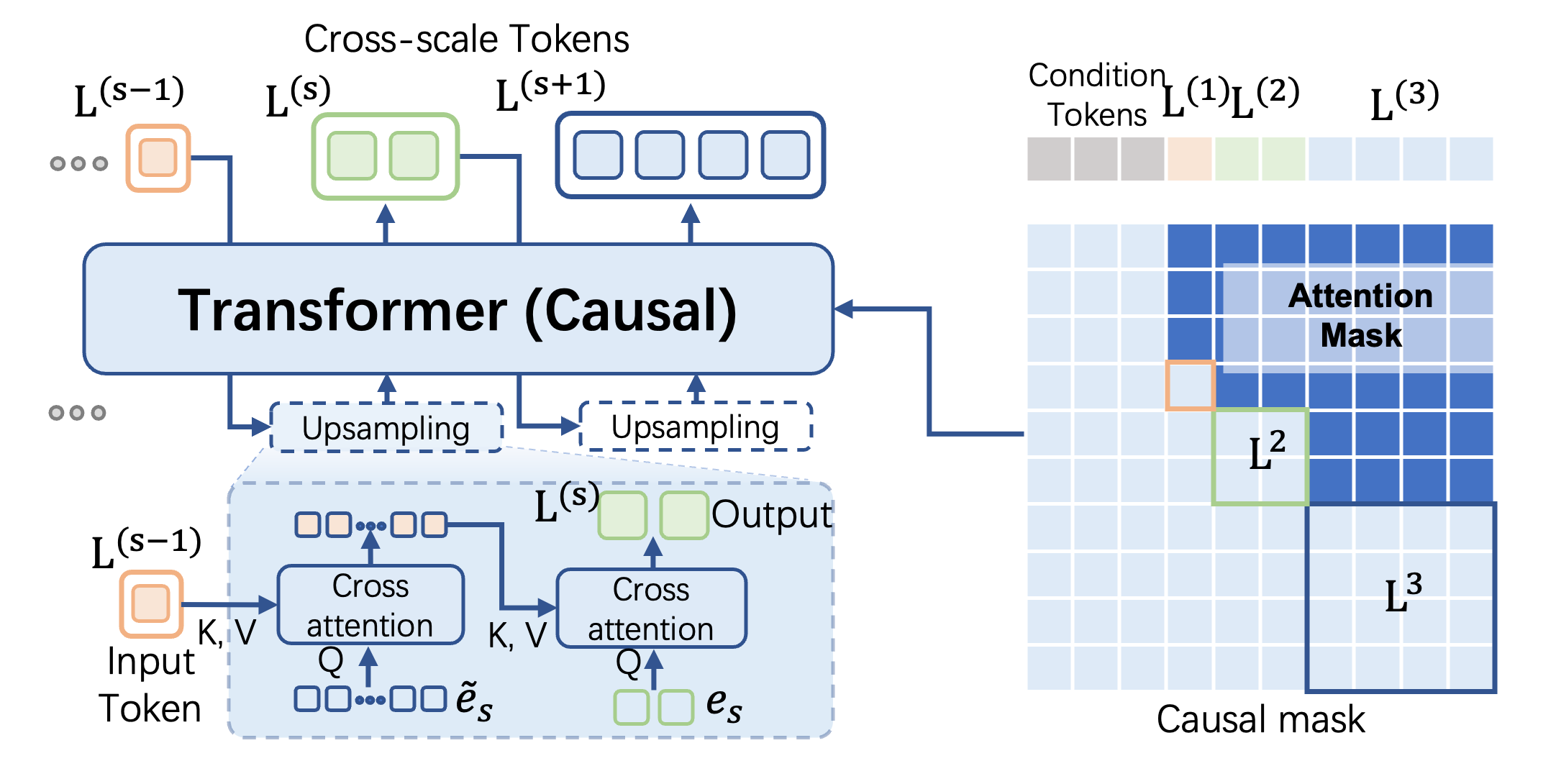} 
\caption{
In next-scale prediction~\cite{VAR} in G3PT, the transformer predicts the next-scale token map using features derived from the ``upsampled" tokens of the previous scale. The ``upsampling" process involves two layers of cross-attention to align the number of tokens across scales. First, features are ``upsampled" with a learnable query $\tilde{e}_s$, and then ``downsampled" using ``downsampling" queries $e_s$ to match the token number of the next scale. 
A causal mask is applied to maintain the correct order and dependencies across different scales and input conditions, ensuring coherence in the model's predictions.
}
\label{fig: CAR}
\end{figure}

\subsection{Overview}



In this section, we introduce our key innovation, the Cross-scale Querying Transformer (CQT), which efficiently maps unordered point-based 3D data into discrete tokens at various levels of detail, creating a sequential relationship ideal for next-scale autoregressive modeling. CQT employs a transformer-based tokenizer to encode high-resolution point clouds into latent feature maps and decode them into 3D occupancy grids through querying points. During quantization, learnable queries compress latent features into discrete tokens at various scales, forming level-of-detail representations that are then decoded back to the original scale using a cross-attention layer.

We first review the key components of tokenization and autoregressive modeling, followed by a detailed description of CQT, including Cross-scale Vector Quantization (CVQ) and Cross-scale Autoregressive Modeling (CAR).

\subsection{Preliminary}

\subsubsection{Tokenization.}

We use Lookup-Free Quantization (LFQ) ~\cite{LFQ} to tokenize the feature map $Z \in \mathbb{R}^{L \times C}$ with $L$ tokens and $C$-dimensional embeddings into the quantized feature map $\hat{Z}$. LFQ streamlines the quantization process by eliminating the need for explicit codebook lookups, thereby reducing the embedding dimension of the feature $Z$. Formally, the quantization is executed via a mapping function $\zeta = q(z)$, which maps a feature vector $z \in \mathbb{R}^{C}$ to an index vector $\zeta \in \mathbb{R}^{\log_2 C}$, with each dimension of $\zeta$ being quantized independently. The token index for $q(z)$ using LFQ is determined by:
\begin{equation}
    \text{Index}(z) = \sum_{i=1}^{\log_2 C} 2^{i-1} \{\zeta_i > 0\}, 
\end{equation}
and the dequantized feature $\hat{z}$ is obtained by:

\begin{equation}
\hat{z} = \hat{q}(\text{sign}(\zeta)) = \hat{q}(\text{sign}(q(z)). \label{eq:quantization}
\end{equation}


\subsubsection{Autoregressive Modeling.}

\textbf{}

\textit{Next-token Prediction.}
For a sequence of discrete tokens $x = (x_1, x_2, \dots, x_N)$, the probability distribution over the sequence is defined as the product of the conditional probabilities of each token given its predecessors, expressed as $P(x) = \prod_{i=1}^N P(x_i \mid x_1, x_2, \dots, x_{i-1})$. This approach effectively models the dependencies between tokens, which is crucial in generating coherent sequences in tasks like language processing and image synthesis. However, in the task of 3D generation, traditional next-token prediction faces significant challenges, as there is no explicit order in the $N$ tokens. 

\textit{Next-scale Prediction.}
Next-scale prediction ~\cite{VAR} focuses on progressively refining the data representation across different scales. This approach involves modeling the probability distribution $P(x)$ over a sequence of multiscale representations, where each level $x^{(s)}$ is conditioned on the preceding coarser scale $x^{(s-1)}$. 
The mathematical formulation is given by:
\begin{equation}
    P(x) = \prod_{s=1}^S P(x^{(s)} \mid x^{(1)}, x^{(2)}, \dots, x^{(s-1)}).
\end{equation}
Here, $x^{(s)} \in \mathbb{R}^{l^{(s)}}$ represents the token ID at the $s$-th scale, where $l^{(s)}$ is the number of tokens. Each scale $x^{(s)}$ contains more detailed information than the previous one, allowing the model to progressively refine the data from a rough approximation to a detailed representation.

However, the original implementation in VAR~\cite{VAR} involves quantizing a feature map 
into $S$ multi-scale token maps $(r_1, r_2, \dots, r_S)$, each at increasingly higher resolutions using average pooling and bilinear interpolation from and to the original scale of the feature map. Therefore, this method imposes an inherent order on the tokens at each scale, which is well-suited for image data but presents challenges for 3D representation. 


\subsection{Cross-scale Querying Transformer}
CQT leverages cross-attention mechanisms to handle unordered tokens through global attention. This approach decomposes unordered 3D point-based features $Z$ (Eq.~\ref{eq:encoder}) into a series of discrete token maps $(r_1, r_2, \dots, r_S)$, each with varying lengths $(L^{(1)}, L^{(2)}, \dots, L^{(S)})$, effectively capturing cross-scale structural information.

\subsubsection{Cross-scale Vector Quantization (CVQ).}

As illustrated in Fig.~\ref{fig: pipeline} (a), we follow the architecture described in 3DShape2VecSet~\cite{3DShape2VecSet} to firstly encode the 3D data. 
{The input point cloud is represented as $X \in \mathbb{R}^{N \times (3+3)}$, with each of the $N$ points having 3 position and 3 normal point features. We employ a cross-attention mechanism to integrate the 3D information from $X$ into the learnable latent queries ${Lat} \in \mathbb{R}^{L \times C}$, as follows:
\begin{equation}
Z = \text{CrossAttn}({Lat}, \text{PosEmb}(X)), \label{eq:encoder}
\end{equation}
where $\text{PosEmb}$ represents Fourier positional encoding and $Z \in \mathbb{R}^{L \times C}$ is the output latent features.
}

{
The quantization process follows the residual quantization steps ~\cite{VAR} with a set of downsample and upsample queries. 
As illustrated in Fig.~\ref{fig: CVQ}(a), conventional downsampling and upsampling methods, such as average pooling and bilinear interpolation, require a sequential arrangement of tokens, which is unsuitable for unordered 3D tokens.
To address this issue, as shown in Fig.~\ref{fig: CVQ}(b), a set of cross-scale learnable queries ${e}_s$ with fewer tokens and one layer of cross-attention are applied to firstly ``downsample'' the unordered tokens. 
Specifically, at each scale, the residual feature $Z_s \in \mathbb{R}^{L \times C}$, which is initialized by $Z_s = Z (s=0)$, is ``downsampled'' into a latent space $E_{s} \in \mathbb{R}^{L^{(s)} \times C}$ with fewer tokens using learnable queries $e_s \in \mathbb{R}^{L^{(s)} \times C}$ and one layer of cross-attention:
\begin{equation}
    E_{s} = {\text{CrossAttn}_{down}}(e_s, Z_s). 
\end{equation}
In this configuration, $e_s$ serves as the query head, while $Z_s$ acts as the key and value heads.

As shonw in Fig.~\ref{fig: CVQ}(b), following the quantization process, denoted as $\hat{E}_{s} = \hat{q}(\text{sign}(q(E_{s})))$, where $\hat{E}_{s}$ is the quantized token given $E_{s}$ using LFQ as illustrated in Eq.~\ref{eq:quantization}, 
the ``upsampled" feature $\hat{Z}_s \in \mathbb{R}^{L \times C}$ at level $s$ is retrieved using another cross-attention layer with an ``upsampled" learnable query $\tilde{e}_s$:
\begin{equation}
    \tilde{Z}_{s} = {\text{CrossAttn}_{up}}(\tilde{e}_s, \hat{E}_{s}), \tilde{e}_s \in \mathbb{R}^{L \times C}. 
\end{equation}

The feature for the subsequent scale, $Z_{s+1}$, is then calculated as: $Z_{s+1} = Z_s - \tilde{Z}_s$. This process iteratively continues until the final quantization step.
The decoder, as shown on the right side of Fig.~\ref{fig: pipeline}(a), which consists of several self-attention layers and a cross-attention layer, decodes these latent codes and a set of query points $p$ into occupancy values:
{\begin{equation}
Occ(p) = \text{CrossAttn}(\text{PosEmb}(p), \text{SelfAttn}(\sum_{s=1}^{S} \tilde{Z}_{s})).
\end{equation}}




}

\subsubsection{Cross-scale AutoRegressive Modeling (CAR).}

{After the CVQ phase, we've obtained cross-scale token maps of varying lengths, serving as inputs for the CAR phase. For the next-scale prediction in CAR, we need to align the token dimensions across scales. Specifically, we employ a cross-scale query transformer, similar to CVQ, to introduce ``upsample" queries that elevate tokens to the latent feature  $\hat{Z}_s \in \mathbb{R}^{L \times C}$ dimension, followed by ``downsample" queries that reduce them to the dimension of the scale to be predicted next.
The initialization of the cross-attention layer's parameters draws from those of the analogous layer in the CVQ phase, with concurrent training alongside the autoregressive transformer to guarantee the correct decompression of the unordered token maps.}

\begin{figure*}[t]
\centering
\includegraphics[width=0.75\textwidth]{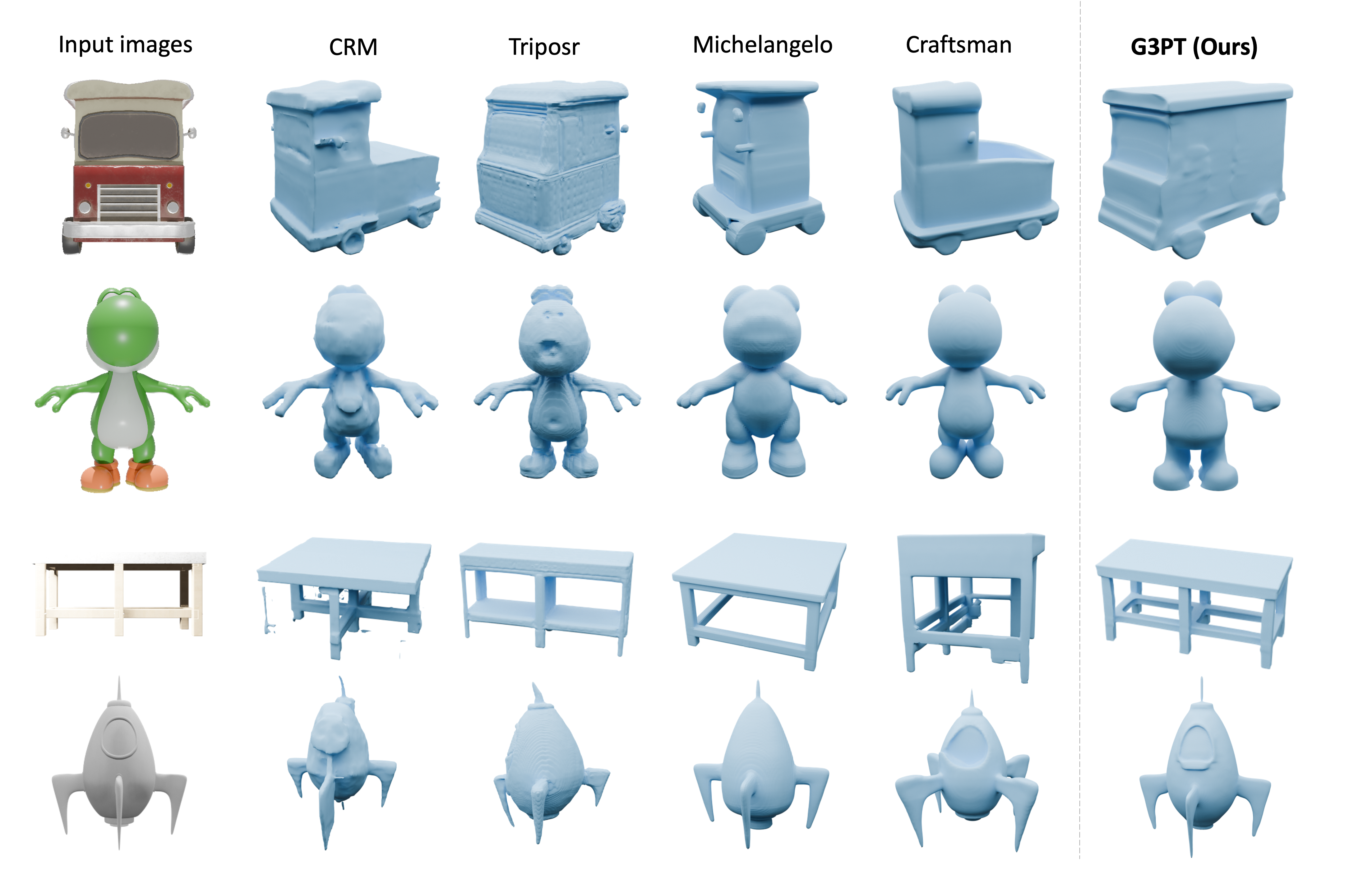} 
\caption{Qualitative comparisons with state-of-the-art methods on the Objaverse dataset~\cite{deitke2023objaverse}.}
\label{fig: qualitative}
\vspace{-1.0em}
\end{figure*}

\section{G3PT: Scaling 3D Generative Model}

\subsection{Conditional Autoregressive Modeling}

In each AR block, pixel-level information from the image and point-level information from the point cloud are seamlessly integrated, aligning the image feature space with the latent space to generate 3D assets consistent with the conditioned content. The overall framework of the AR model is depicted in Fig.~\ref{fig: pipeline}(b), while the architecture of each AR block is illustrated in Fig.~\ref{fig: CAR}.

Specifically, the pre-trained DINO-v2 (ViT-L/14)~\cite{oquab2023dinov2} is employed to extract image-conditioned tokens, leveraging its strength in capturing the structural information crucial for 3D tasks. A linear layer projects the $N_I$ image tokens $I_{dino} \in \mathbb{R}^{L_I \times C_I}$, derived from DINO-v2, to match the channel dimension of the cross-scale 3D tokens $Z_{dino} \in \mathbb{R}^{L_I \times C}$. 
These image tokens are then concatenated with the cross-scale 3D tokens and regulated through an attention mask in the causal transformer, ensuring that only subsequent 3D tokens are predicted.


It is important to note that this approach represents a basic attempt to validate the model's flexible conditioning capabilities; there are numerous other conditioning methods that warrant further exploration.

Text condition can also be easily added to ensure semantic consistency using the pre-trained CLIP model, extracts semantic tokens $c_s$ from the conditional text input. Additionally, adaLN~\cite{wu2024direct3d} is used to control the signal.

\subsection{Training Details}

Directly training with a large number of discrete tokens is highly time-consuming. To mitigate this, the initial training phase for the CQT involves training an encoder-decoder model without quantization, incorporating layer normalization between the encoder and decoder. Once this model is adequately trained, the parameter in CQT  is used to fine-tune the quantization layer.
For training the CAR component, a progressive training method is implemented. Instead of processing all tokens across all scales $(r_1, r_2, \dots, r_S)$ simultaneously, the training begins with tokens before the $S/2$ scale $(r_1, r_2, \dots, r_{S/2})$ and progressively includes larger scales. This approach accelerates convergence and significantly improves training stability.

\section{Experiments}

\subsection{Implementation Details}

The input of CQT consists of point clouds, each containing 16384 points uniformly sampled from the 3D model in the Objaverse dataset~\cite{deitke2023objaverse}, accompanied by a learnable query with length $L=2304$ and channel dimension $C = 512$. The vocabulary size of the codebook in LFQ is 8192. The encoder network includes one cross-attention layer and 12 self-attention layers. The decoder network contains 16 self-attention layers with the same channel dimension of the encoder.  During training, 8192 uniform points and 8192 near-surface points are sampled for supervision. The AdamW optimizer is employed with a learning rate of $1 \times 10^{-4}$, and the model is trained for 60,000 steps on 8 NVIDIA A100 GPUs with 80GB memory.

The CAR transformer adopts the architecture similar to the standard decoder-only transformers like GPT-2. 
To stabilize training, queries and keys are normalized to unit vectors before attention is applied. 
All models are trained with the same learning rates and batch sizes, with a learning rate of $1 \times 10^{-5}$ using the AdamW optimizer with $\beta_1 = 0.9$, $\beta_2 = 0.95$, and a weight decay of 0.05 for every 1000 steps. The 1.5B model is trained for two weeks on 136 NVIDIA H20 GPUs with 96GB memory.

\subsection{State-of-the-art 3D Generation and Scaling Behaviors}

\begin{figure*}[t]
\centering
\includegraphics[width=0.7\textwidth]{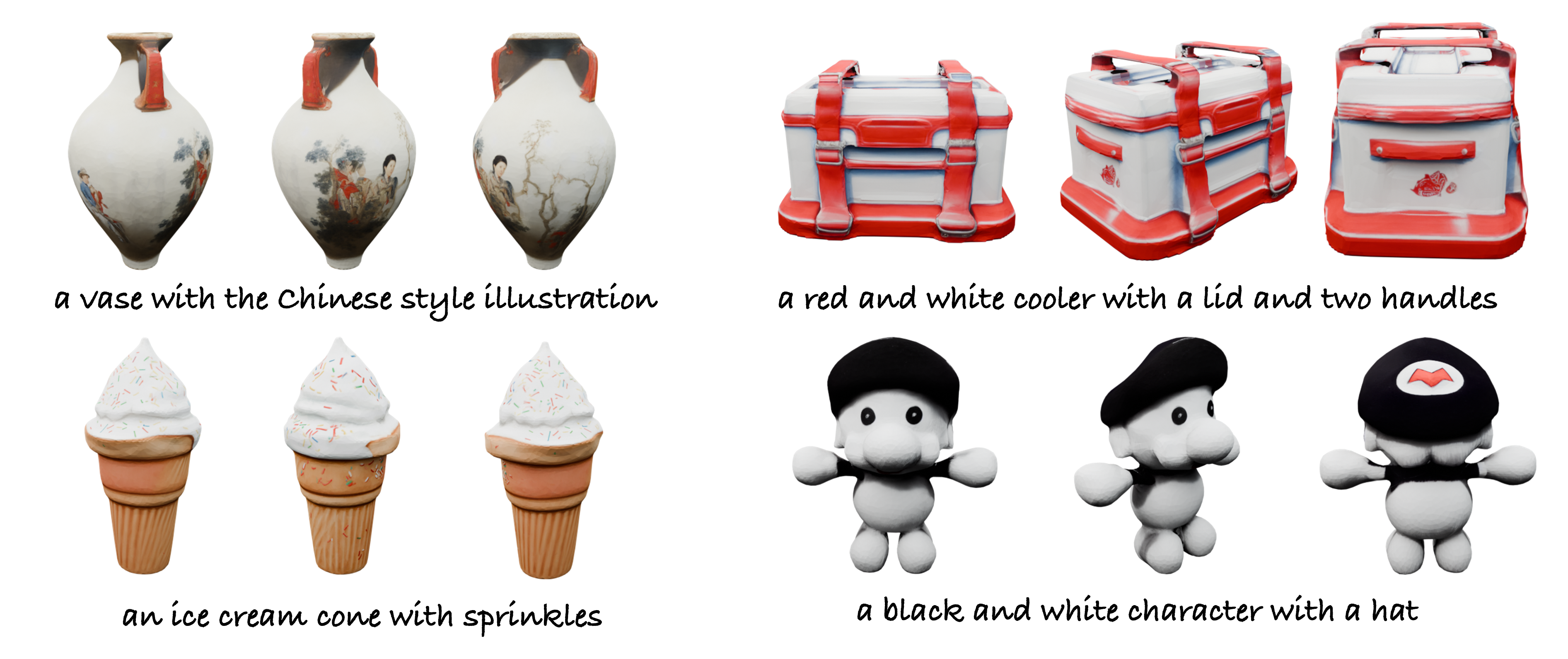} 
\caption{Mesh visualization using SyncMVD~\cite{liu2023text} to generate textures for the meshes
produced by G3PT.}
\label{fig: texture}
\vspace{-1.0em}
\end{figure*}


\begin{table}[h]
\centering
\setlength{\tabcolsep}{1.5pt}
\begin{tabular}{ccc|ccc}
\toprule
Type & Method & Name & IoU$\uparrow$ & Cham.$\downarrow$ & F-score$\uparrow$ \\
\midrule
\multirow{4}{*}{LRM} & \multirow{3}{*}{NeRF} & Triposr & 72.6 & 0.023 & 58.2 \\
 & & InstantMesh & 68.7 & 0.029 & 58.3 \\
 & & CRM & 76.3 & 0.020 & 61.4 \\
 & Gaussian & LGM & 67.6 & 0.025 & 49.3 \\
\midrule
\multirow{7}{*}{\begin{tabular}[c]{@{}c@{}}3D\\Generation\end{tabular}} 
& \multirow{4}{*}{Diffusion} & Michelangelo & 74.5 & 0.028 & 62.5 \\
 & & Shap-e & 66.8 & 0.029 & 46.3 \\
 & & CraftsMan & 72.2 & 0.021 & 56.1 \\
 & & CLAY*(0.5B) & 77.1& 0.021& 63.4 \\
 
\cline{2-6}

& \multirow{3}{*}{\begin{tabular}[c]{@{}c@{}}AR\\Modeling\end{tabular}}
& G3PT(0.1B) & 73.9 & 0.025 & 60.4 \\
&  & G3PT(0.5B) & 82.11 & 0.015 & 75.1\\
& & \textbf{G3PT(1.5B)} & \textbf{87.6}& \textbf{0.013}& \textbf{83.0} \\
\bottomrule
\end{tabular}
\caption{Comparison of state-of-the-art 3D generation methods. (*: Reproduction)}
\label{tab:comparison}
\end{table}

The quantitative comparisons are presented in Table~\ref{tab:comparison}. The evaluated methods include LRM-based approaches such as InstantMesh~\cite{xu2024instantmesh} and CRM~\cite{wang2024crm}, which integrates Flexicubes~\cite{shen2023flexible} into large reconstruction models to optimize mesh representations for enhanced smoothness and geometric detail. Triposr~\cite{tochilkin2024triposr} utilizes a transformer backbone to map image tokens to implicit 3D triplanes under multi-view image supervision, while LGM~\cite{tang2024lgm} replaces the triplane NeRF representation with 3D Gaussians~\cite{kerbl20233d} to improve rendering efficiency.
Additionally, diffusion-based methods such as Michelangelo~\cite{zhao2024michelangelo}, Shap-E~\cite{jun2023shap}, CraftsMan~\cite{li2024craftsman}, and CLAY~\cite{zhang2024clay} are compared. These methods first sample point clouds from the mesh and then employ a transformer-based Variational Autoencoder~(VAE) to encode the input point cloud into an implicit latent space, from which a latent diffusion model is trained to generate 3D shapes. 

The reconstructed mesh quality is evaluated using Intersection Over Union (IoU), Chamfer distance (Cham.)—where lower values indicate better performance—and F-score (with a threshold of 0.01), which together reflect the proximity of the reconstructed mesh to the GT mesh.
The results highlight the significant advantage of G3PT, particularly the model with 1.5 billion parameters, which outperforms all other methods with a substantial margin in all metrics, demonstrating superior generation quality and fidelity.

As shown in Fig.~\ref{fig: qualitative}, qualitative comparisons of G3PT with other state-of-the-art methods are performed on the Objaverse dataset for the image-to-3D task. The comparison includes LRM-based methods such as CRM~\cite{wang2024crm} and Triposr~\cite{tochilkin2024triposr}, as well as diffusion-based methods like Michelangelo~\cite{zhao2024michelangelo} and CraftsMan~\cite{li2024craftsman}.
LRM-based methods generate 3D models that more closely resemble the input images but often suffer from noise and mesh artifacts.
Diffusion-based methods like Michelangelo produce plausible geometry but struggle to align with the semantic content of the conditional images. 
In contrast, as shown in Fig.~\ref{fig: qualitative}, G3PT achieves a superior balance between quality and realism, consistently producing high-quality meshes that align well with the conditional images in most cases. 
Moreover, G3PT generates smooth and intricate geometric details, enhancing the effectiveness of existing texture synthesis techniques. This high-quality meshes produced by G3PT allow for easy texture generation using SyncMVD~\cite{liu2023text}, as demonstrated in Fig.~\ref{fig: texture}.

\begin{figure}[t]
\centering
\includegraphics[width=0.45\textwidth]{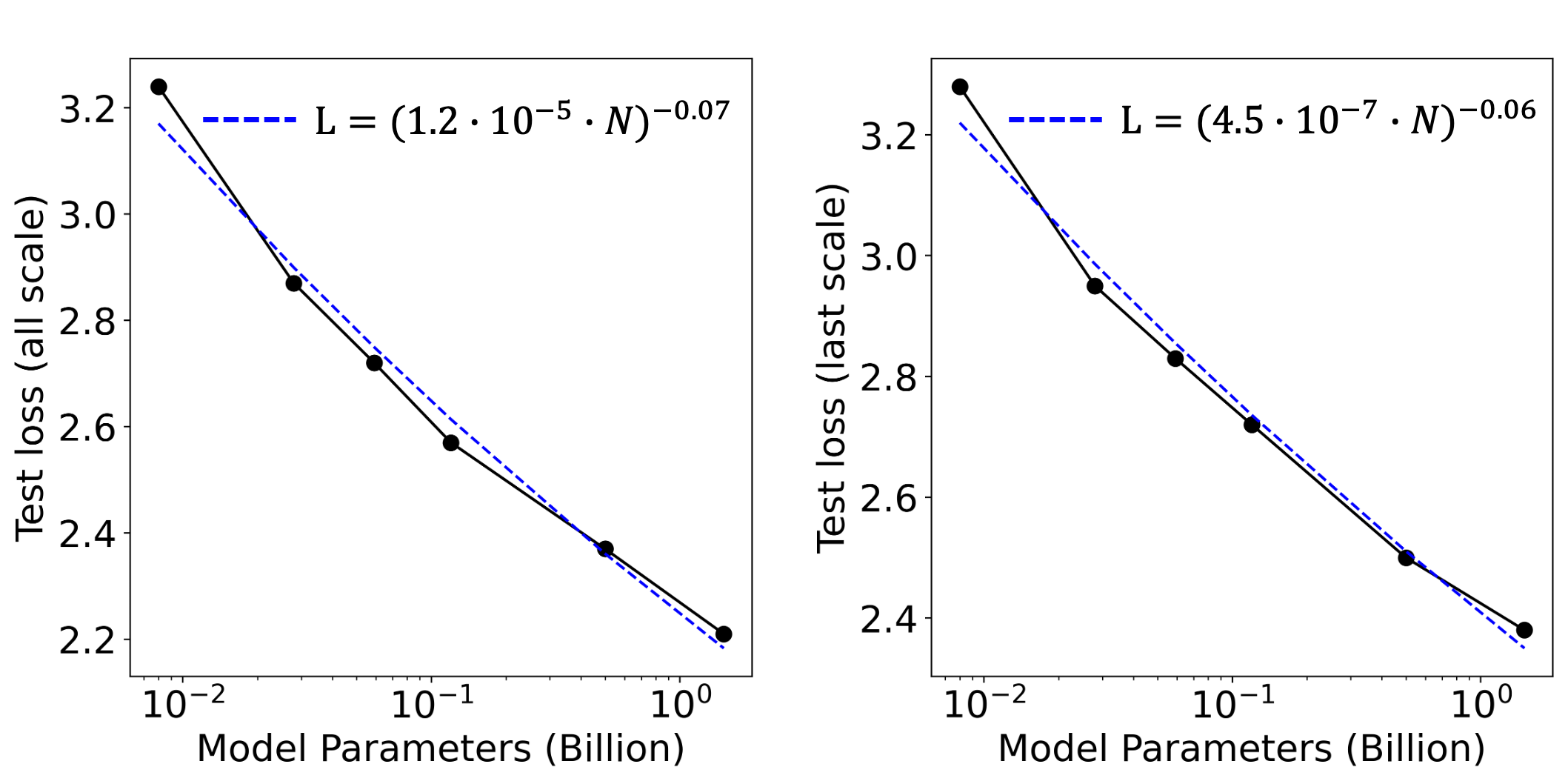} 
\caption{Scaling laws in G3PT with network parameters $N$.}
\vspace{-8pt}
\label{fig: scaling}
\end{figure}

Furthermore, Fig.~\ref{fig: scaling} illustrates the scaling laws observed in G3PT by examining the relationship between model parameters (in billions) and test loss (measured as cross-entropy). Both plots demonstrate a clear trend where the test loss decreases as the number of model parameters increases with a power-law relationship. This indicates that as the model size grows, the performance improves, further validating the potential of G3PT in handling complex 3D generation tasks.

\begin{table}[htb]
    \centering
    \setlength{\tabcolsep}{1pt}
    \begin{tabular}{cc|ccccc}
        \toprule
        Method & \#Token & IOU$\uparrow$ & Cham.$\downarrow$ & F-score$\uparrow$ & Acc.(\%)$\uparrow$ & Usage(\%)$\uparrow$ \\
        \midrule
        \multirow{2}{*}{VAE} & 576 & 89.20 & 0.0126 & 84.10 & 95.24 & - \\
                             & 2408 & 89.60 & 0.0118 & 85.80 & 95.80 & - \\
        \midrule
        \multirow{2}{*}{VQVAE} & 576 & 85.32 & 0.0134 & 80.15 & 85.59 & 96.34 \\
                               & 2408 & 87.43 & 0.0131 & 80.53 & 88.32 & 92.96 \\
        \midrule
        \multirow{2}{*}{CQT} & 576 & 89.38 & 0.0122 & 85.70 & 95.27 & \textbf{99.51} \\
                                & 2408 & \textbf{90.35} & \textbf{0.0108} & \textbf{87.23} & \textbf{97.13} & 97.13 \\
        \bottomrule
    \end{tabular}
    \caption{Performance comparison of different tokenizers.}
    \label{tab:cqt_comparison}
\end{table}

\begin{table*}[t]
    \centering
    \begin{tabular}{cc}
        \begin{subtable}[t]{0.48\textwidth}
            \centering
            \setlength{\tabcolsep}{2pt}
            \begin{tabular}{cc|ccccc}
                \toprule
                \#Token & \begin{tabular}[c]{@{}c@{}}Codebook \\ Size \end{tabular} & IOU(\%) & Cham. & F-score(\%) & Acc.(\%) \\
                \midrule
                2408 & 1024 & 85.43& 0.0122& 81.25& 92.96\\
                2408 & 2048 & 86.65& 0.0115& 86.36& 94.44\\
                2408 & 4096 & 89.32 & 0.0119 & 87.46 & 96.21 \\
                2408 & \textbf{8192} & \textbf{90.35} & \textbf{0.0108} & \textbf{87.23} & \textbf{97.13} \\
                \bottomrule
            \end{tabular}
            \caption{Codebook sizes. }
            \label{tab:performance_metrics_1}
        \end{subtable} &
        \begin{subtable}[t]{0.48\textwidth}
            \centering
            \setlength{\tabcolsep}{2pt}
            \begin{tabular}{cc|ccccc}
                \toprule
                \#Token & \begin{tabular}[c]{@{}c@{}}Codebook \\ Size \end{tabular} & IOU(\%) & Cham. & F-score(\%) & Acc.(\%) \\
                \midrule
                256 & 8192 & 84.31 &0.0125& 80.18& 93.32 \\
                576 & 8192 & 89.38& 0.0122& 85.7 &95.27 \\
                1024 & 8192 & 89.51 & 0.0119 & 86.86 & 96.58 \\
                \textbf{2408} & 8192 & \textbf{90.35} & \textbf{0.0108} & \textbf{87.23} & \textbf{97.13} \\
                \bottomrule
            \end{tabular}
            \caption{Number of tokens.}
            \label{tab:performance_metrics_2}
        \end{subtable}
    \end{tabular}\caption{Performance of different number of tokens and codebook sizes.}\label{tab:size_token}
\end{table*}

\begin{table*}[h]
\centering
\setlength{\tabcolsep}{3pt}
\begin{tabular}{ccc|ccccc}
 \toprule
\begin{tabular}[c]{@{}c@{}}3D \\ Representation \end{tabular}  & \begin{tabular}[c]{@{}c@{}}Encoding \\ Method \end{tabular} & Quantization & IOU (\%)$\uparrow$ & Cham.$\downarrow$ & F-score (\%)$\uparrow$ & Acc. (\%)$\uparrow$ & Usage (\%)$\uparrow$ \\
\midrule
Volume & 3D CNN & LFQ & 85.29 & 0.0114 & 78.66 & 86.89 & 91.34 \\
Triplane & Learnable query & LFQ & 86.13&  0.0120 & 80.12 & 90.44& 89.34 \\
 \multirow{2}{*}{1D Latent} 
  & Learnable query & Pooling + LFQ & 89.51 & 0.0139 & 86.86 & 93.80 & \textbf{99.50} \\
 & Learnable query & CVQ & \textbf{90.35} & \textbf{0.0108} & \textbf{87.23} & \textbf{97.13} & 97.13 \\
 
\bottomrule
\end{tabular}
\caption{Comparison of different encoding methods across various 3D representations.}
\vspace{-15pt}
\label{tab:representation}
\end{table*}

\subsection{Tokenization}

This section demonstrates the state-of-the-art performance of the Cross-scale Query Transformer (CQT) in comparison to other tokenization methods. A total of 1200 models are randomly selected for evaluation. Reconstruction quality is assessed by geometric accuracy, specifically comparing the reconstructed mesh to the ground truth (GT) mesh.


Table~\ref{tab:cqt_comparison} presents a comparison of three methods: VAE, VQVAE, and CQT. VAE follows the approach outlined by Zhang et al.~\cite{zhang2024clay}, which applies KL regularization between the encoder and decoder, differing from the quantization module utilized in diffusion-based model training. VQVAE incorporates LFQ quantization, maintaining the same quantization structure as described by~\cite{LFQ}.
Performance metrics include prediction accuracy (Acc.) of the occupancy value (0 or 1), which is determined by evaluating points that are randomly sampled in the vicinity of the ground truth (GT) mesh. 
The ``Usage" value indicates the efficiency of codebook usage by each tokenizer.
CQT outperforms both VAE and VQVAE across multiple metrics with near-complete codebook usage. 


A detailed comparison of the performance metrics across different codebook sizes and token numbers are presented in Table ~\ref{tab:size_token}. 
Due to memory constraints, the evaluation was limited to a maximum of 8192 tokens. 
Similarly, Table~\ref{tab:performance_metrics_2} explores the impact of different token numbers while keeping the codebook size fixed at 8192. 
The comparison of various 3D representation and encoder architectures is shown in Table~\ref{tab:representation}, which considers three types of 3D representations: volumetric feature grids (Volume), Triplane~\cite{wang2023petneus}, and 1D latent vectors. Each representation is paired with a specific encoder method. The Volume representation is encoded using a 3D Convolutional Neural Network (3D CNN) following the architecture of SDFusion~\cite{cheng2023sdfusion}, while the Triplane representation utilizes the architecture from Direct3D~\cite{wu2024direct3d}. All encoders are designed with a similar number of parameters.

A comparison is also made between the proposed CVQ and a baseline implementation without the Cross-scale Querying Transformer. In the baseline setup, a similar implementation to VAR~\cite{VAR} is used, incorporating an 1D average pooling module, a bilinear upsampling module (Fig.~\ref{fig: CVQ}(a)), and an additional 1D convolutional layer, which forces the tokens to learn an ordered sequence.

The results demonstrate that the CVQ method achieves superior performance 
with near-complete codebook usage. This emphasizes the effectiveness of CVQ in preserving detailed structural information during quantization.
Additionally, when comparing CVQ to the baseline method using average pooling, the results show that average pooling reduces reconstruction accuracy. This reduction is attributed to the fact that 1D latent tokens do not possess an inherent sequential order, unlike image data. 

\section{Conclusion}

This paper introduces G3PT, a scalable coarse-to-fine 3D generative model that leverages a Cross-scale Querying Transformer (CQT) to effectively map unordered 3D data into discrete tokens across various levels of detail. By establishing a natural sequential relationship among these tokens, G3PT enables autoregressive modeling in a manner that aligns well with the inherent characteristics of 3D data. This innovative cross-scale autoregressive framework tailored for unordered data offers new insights into autoregressive algorithm design. Extensive experiments demonstrated that G3PT achieves superior generation ability compared to existing 3D generation methods, setting a new state-of-the-art in 3D content creation.

\subsubsection{Future works.}
The current model requires substantial computational resources, and the training process is time-intensive. Future work could explore more efficient training techniques and investigate additional conditioning methods.



\bibliography{aaai25}


\end{document}